\documentclass[conference]{IEEEtran}
\IEEEoverridecommandlockouts
\usepackage{xcolor,colortbl}
\usepackage{array}
\usepackage{multirow}
\usepackage{graphicx}
\usepackage{booktabs}
\usepackage{cite}
\usepackage{amsmath,amssymb,amsfonts}
\usepackage{algorithmic}
\usepackage{graphicx}
\usepackage{textcomp}
\usepackage{xcolor}
\def\BibTeX{{\rm B\kern-.05em{\sc i\kern-.025em b}\kern-.08em
    T\kern-.1667em\lower.7ex\hbox{E}\kern-.125emX}}
\begin{document}

\title{Quaternion Recurrent Neural Network \\with Real-Time Recurrent Learning \\and Maximum Correntropy Criterion
}

\author{\IEEEauthorblockN{Pauline Bourigault}
\IEEEauthorblockA{\textit{Department of Computing} \\
\textit{Imperial College London}\\
\textit{London, United Kingdom}\\
p.bourigault22@imperial.ac.uk}
\and
\IEEEauthorblockN{Dongpo Xu}
\IEEEauthorblockA{\textit{School of Mathematics and Statistics} \\
\textit{Northeast Normal University}\\
\textit{Changchun, P.R.China}\\
xudp100@nenu.edu.cn}
\and
\IEEEauthorblockN{Danilo P. Mandic}
\IEEEauthorblockA{\textit{Department of Electrical and} \\
\textit{Electronic Engineering}\\
\textit{Imperial College London}\\
\textit{London, United Kingdom}\\
d.mandic@imperial.ac.uk}
}

\maketitle

\begin{abstract}
We develop a robust quaternion recurrent neural network (QRNN) for real-time processing of 3D and 4D data with outliers. This is achieved by combining the real-time recurrent learning (RTRL) algorithm and the maximum correntropy criterion (MCC) as a loss function. While both the mean square error and maximum correntropy criterion are viable cost functions, it is shown that the non-quadratic maximum correntropy loss function is less sensitive to outliers, making it suitable for applications with multidimensional noisy or uncertain data. Both algorithms are derived based on the novel generalised HR (GHR) calculus, which allows for the differentiation of real functions of quaternion variables and offers the product and chain rules, thus enabling elegant and compact derivations. Simulation results in the context of motion prediction of chest internal markers for lung cancer radiotherapy, which includes regular and irregular breathing sequences, support the analysis.
\end{abstract}


\section{Introduction}
\label{sec:intro}
Recently, the incorporation of quaternion algebra into neural network architectures has paved the way for enhancing performance and robustness, particularly in the contexts where data are naturally multidimensional \cite{Isokawa2003,Parcollet2019_reviewQNN,ZhuQuatCNN, QuatKnowlGraphEmbed}. Quaternion neural networks (QNNs) leverage the inherent multidimensional nature of quaternions to build models that are more compact than quadrivariate real ones, thus improving parameter efficiency and potentially capturing intricate patterns in data \cite{Gaudet2018,Parcollet2019,Takahashi2022,Comminiello2019_3D}. Consequently, extensions of the real-valued backpropagation to the domain of quaternions has led to various QNN architectures \cite{Xu2016_Optimization}. Notably, the development of learning algorithms using the Generalised HR (GHR) calculus has been shown to enable new algorithms that make full use of the structurally rich quaternion algebra and so enhance performance \cite{Xu2015_GHR,Xu2017,Popa2017}.

In the context of recurrent neural networks (RNNs), the real-time recurrent learning (RTRL) algorithm, owing to its online nature, is a preferred choice for real-time applications \cite{Williams1989,Mandic2001}. It is therefore natural to investigate whether, in conjunction with quaternions, the RTRL would yield similar advantages when tracking non-linear dynamics and adapting to intricate temporal patterns in multidimensional sequences.

Current QNNs are mostly trained with the mean square error (MSE) cost function \cite{Parcollet2018QRNN}. Since the MSE measures the average squared difference between the predicted and actual data values, it is sensitive to outliers and is optimal only for normally distributed data. Unlike MSE, the maximum correntropy criterion (MCC) is a non-quadratic loss function, which employs a nonlinear kernel to measure the similarity between the actual and predicted data \cite{Santamaria2006}. This makes the MCC less sensitive to outliers, and hence suitable for applications with noisy and heavy tailed data. For quaternion recurrent neural networks (QRNNs), where data are corrupted with multidimensional noise of different channel-wise natures, the MCC promises to offer a robust alternative to MSE.

This motivates us to introduce a QRNN trained with RTRL and the MCC cost in this setting. The novel GHR calculus is employed for compact derivations of otherwise very cumbersome gradient expression in quaternion learning machines. The performance of the proposed network is compared against its counterpart that employs MSE as a loss function, and against the RNN equipped with RTRL and MCC or MSE loss, the Quaternion Least Mean Square (QLMS) and the Least Mean Square (LMS) algorithms. The performances are verified in the context of motion prediction of chest internal markers for lung cancer radiotherapy. The approach is general enough to serve as a basis of a whole class of online QNNs.

\section{Preliminaries}
\label{sec:preliminaries}
\subsection{Quaternion algebra}
Denote a quaternion, \textit{q}, as 
\begin{equation}
    q = q_a + iq_b + jq_c + kq_d \,,
\end{equation}
where $q_a , q_b, q_c, q_d \in \mathbb{R}$ and the imaginary units \textit{i}, \textit{j}, and \textit{k} satisfy $i^2 = j^2 = k^2 = ijk = -1$, $ij = -ji = k$, $jk = -kj = i$, $ki = -ik = j$. The set of quaternions is defined as $\mathbb{H} \triangleq \{q = q_a + iq_b + jq_c +kq_d \; | \; q_a, q_b, q_c, q_d \in \mathbb{R}\} $. The multiplication of two quaternions in $\mathbb{H}$ is noncommutative due to the properties of the imaginary units. The real part of $q$ is defined as $\operatorname{Re}\{q\} = q_a$, whereas the imaginary part is $\operatorname{Im}\{q\} = iq_b + jq_c + kq_d$. The conjugate of $q$ is $q^* = \operatorname{Re}\{q\} - \operatorname{Im}\{q\} = q_a - iq_b - jq_c - kq_d$. The modulus of a quaternion is denoted as $|q| = \sqrt{qq^*}$. The inverse of a quaternion $ q \neq 0$ is $q^{-1} = q^* / |q|^2$. For any quaternion, \textit{q}, and a nonzero quaternion, $\mu$, the transformation \cite{Ward1997}
\begin{equation}
    q^{\mu} \triangleq \mu  q \mu^{-1}
\end{equation} 
describes a rotation of \textit{q}. 

\subsection{The GHR calculus}
A quaternion function $ f : \mathbb{H} \rightarrow \mathbb{H}$, defined as $f(q) = f_a(q_a,q_b,$ $q_c,q_d) + if_b(q_a,q_b,q_c,q_d) + jf_c(q_a,q_b,q_c,q_d) + kf_d(q_a,q_b,q_c,$ $q_d)$ is called real-differentiable, if $f_a,\, f_b,\,f_c,\,f_d$ are differentiable as functions of the real variables $q_a,\,q_b,\,q_c,\,q_d$ \cite{Sudbery1979}. If $ f : \mathbb{H} \rightarrow \mathbb{H}$ is real-differentiable, then the left GHR derivative of the function $f$ with respect to $q^{\, \mu}$ ($\mu \neq 0, \mu \in \mathbb{H}$) is
\begin{equation}
    \frac{\partial f}{\partial q^\mu}  = \frac{1}{4} \biggl(\frac{\partial f}{\partial q_a} - \frac{\partial f}{\partial q_b}i^\mu - \frac{\partial f}{\partial q_c}j^\mu - \frac{\partial f}{\partial q_d}k^\mu   \biggr) \in \mathbb{H}\, ,
\end{equation}
where $q = q_a + iq_b + jq_c + kq_d$, $q_a,q_b,q_c,q_d \in \mathbb{R}$, and $\frac{\partial f}{\partial q_a}, \frac{\partial f}{\partial q_b}, $ $\frac{\partial f}{\partial q_c}, \frac{\partial f}{\partial q_d} \in \mathbb{H}$ are the partial derivatives of $f$ with respect to $q_a,q_b,q_c,q_d$. If $ f : \mathbb{H} \rightarrow \mathbb{H}\,$, $ g : \mathbb{H} \rightarrow \mathbb{H}\,$, the product rule is \cite{Xu2015_GHR}
\begin{equation}
    \frac{\partial (fg)}{\partial q^\mu} = f \frac{\partial g}{\partial q^\mu} + \frac{\partial f}{\partial q^{g\mu}}g \,,
\end{equation}
the chain rule is
\begin{equation}
    \frac{\partial f(g(q))}{\partial q^\mu} = \sum_{\upsilon \in \{1,i,j,k \}} \frac{\partial f}{\partial g^{\upsilon}} \frac{\partial g^{\upsilon}}{\partial q^{\mu}} \,,
\end{equation}
and the rotation rule is
\begin{equation}
    \biggl( \frac{\partial f}{\partial q^\mu}  \biggr)^\upsilon = \frac{\partial f^\upsilon}{\partial q^{\upsilon \mu}} \, .
\end{equation}
Denote the quaternion gradient of a function $ f : \mathbb{H}^n \rightarrow \mathbb{R}$ as
\begin{equation}
    \nabla_{\textbf{\textit{q}}}f \triangleq \biggl( \frac{\partial f}{\partial \textbf{\textit{q}}}    \biggr)^T = \biggl(  \frac{\partial f}{\partial q_1}, \cdots,  \frac{\partial f}{\partial q_n}  \biggr)^T \in \mathbb{H}^n \, ,
\end{equation}
where $\Bigl( \frac{\partial f}{\partial \textbf{\textit{q}}}    \Bigr)^T$ is the transpose of $\frac{\partial f}{\partial \textbf{\textit{q}}}$ \cite{Xu2016_Optimization}. The quaternion Jacobian matrix of $\textbf{f} : \mathbb{H}^{N \times 1} \rightarrow \mathbb{H}^{M\times1}$ is then defined as
\begin{equation}
  \frac{\partial \textbf{f}}{\partial\textbf{q}}  =   
        \begin{pmatrix}
            \frac{\partial f_1}{\partial q_1} & \cdots & \frac{\partial f_1}{\partial q_N}   \\
            \vdots & \ddots & \vdots   \\
            \frac{\partial f_M}{\partial q_1} & \cdots & \frac{\partial f_M}{\partial q_N}
        \end{pmatrix}.
\end{equation}
Note the convention that for two vectors $\textbf{f} \in \mathbb{H}^{M\times1}$ and $\textbf{q} \in \mathbb{H}^{N \times 1}$, $\frac{\partial \textbf{f}}{\partial \textbf{q}}$ is a matrix for which the $(m, n)$th element
is $(\partial f_m/ \partial q_n)$, thus, the dimension of $\frac{\partial \textbf{f}}{\partial \textbf{q}}$ is $M \times N$ \cite{Xu2016_Optimization}.

\subsection{Maximum Correntropy criterion}
\begin{figure}
    \centering
    \includegraphics[width=\linewidth]{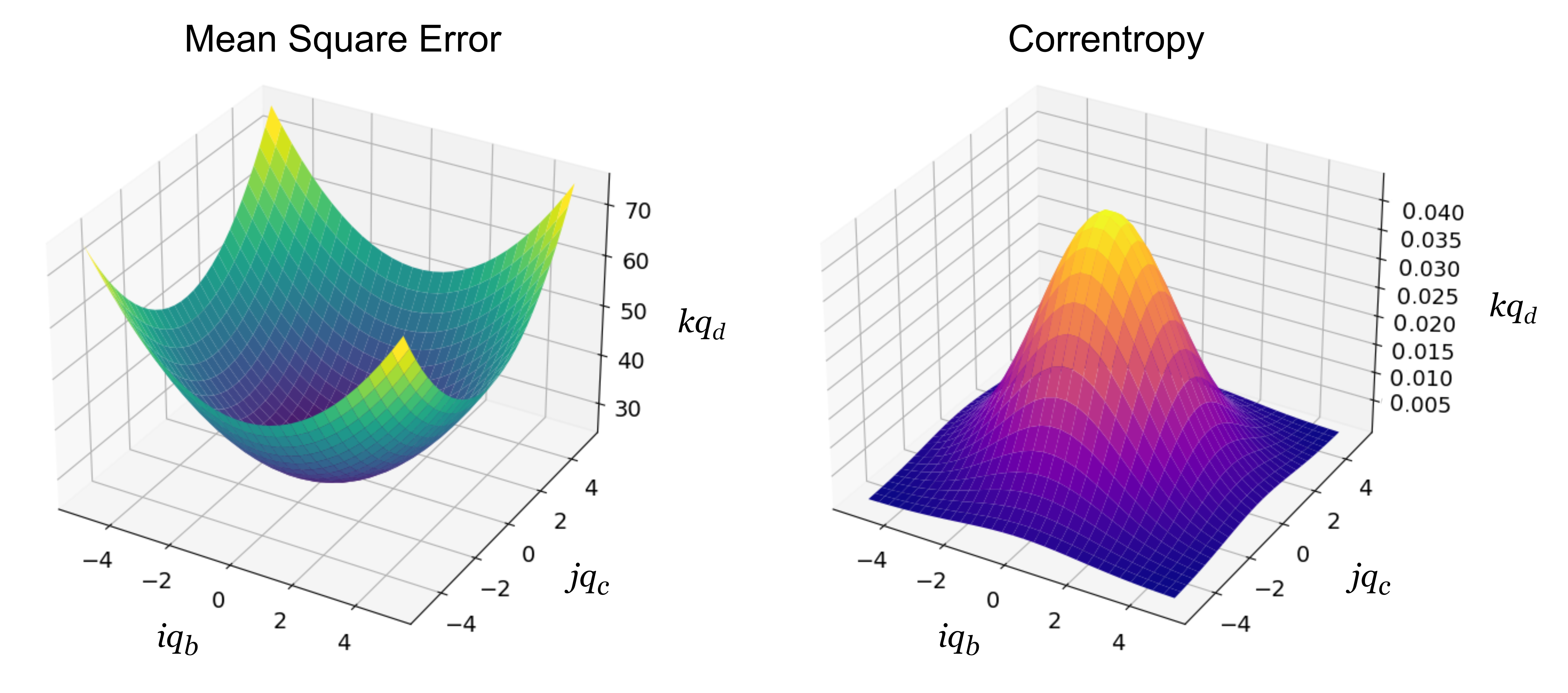}
    \caption{Mean square error (MSE) (left) and correntropy (right) in the 3D space where each point represents a pure quaternion. We assume that the true quaternion is (0, 0i, 0j, 0k) for simplicity.}
    \label{fig:MSE_Corr}
\end{figure}
The aim is to form a quaternion function $z$, based on a sequence $(\textbf{s}_1,d_1), (\textbf{s}_2,d_2), \dots ,(\textbf{s}_N,d_N)$, where $\textbf{s}_p$ is the system input, and $d_p$ is the expected response, both quaternion valued, for an instant time $p$. We denote the expected response as $d_p=z_{opt}(\textbf{s}_p)+\xi_p$, where $z_{opt}$ is a nonlinear quaternion function to estimate, and $\xi_p$ is noise with an arbitrary probability density function that does not need to be Gaussian.

When finding $z$, we ideally would like to solve the empirical risk minimization problem, that is, to minimize 
\begin{equation} \mathrm{R}_{emp}[f\in Q,{\mathbf{Z}}_N]=\sum_{p=1}^N {\left| {z_{opt} ({\mathbf{s}}_p )-z({\mathbf{s}}_p )}\right|^2} 
\end{equation}
where $Q$ is a quaternion vector space. Notably, the use of MSE can result in large variations for the weights, or shift in the output, when the noise distribution contains outlying points, is non-symmetric, or has a nonzero mean. 

The correntropy is defined as $ V_\sigma (X,Y)=E\left[\kappa_\sigma (X - Y)\right]$, where $X$ and $Y$ are quaternion random variables, and $\kappa_\sigma(\cdot)$ is a symmetric positive-definite quaternion kernel, with kernel size $\sigma$ \cite{WeifengLiu2006}. For real-valued inputs, the Gaussian kernel can be expressed as 
\begin{equation}\label{eq:MCC_real}
    \kappa_{real,\sigma}(X-Y)=\frac{1}{\sqrt{2\pi}\sigma}\exp\left[-\frac{1}{2\sigma ^2}(X - Y)^2\right] 
\end{equation}
where $X$ and $Y$ are real valued \cite{Ogunfunmi2015}. 
An analogous Gaussian-based kernel for quaternion data may be expressed as 
\setlength{\arraycolsep}{0.0em}
\begin{eqnarray}\label{eq:MCC_quat_kernel}
\kappa_\sigma (X - Y) &{}={}&\frac{4}{\sqrt {2\pi}\sigma}\exp \biggl[ -\frac{1}{2\sigma^2} \Bigl[ (X_r - Y_r )^2+(X_i - Y_i )^2 \nonumber\\
&&+\:(X_j - Y_j )^2+(X_k - Y_k )^{2} \Bigr] \biggr] \nonumber\\
&=& \frac{4}{\sqrt {2\pi} \sigma}\exp\left[-\frac{{  1}}{2\sigma^2}\left| {X - Y} \right|^2\right]
\end{eqnarray}
where $X$ and $Y$ are the quaternions $X=X_r+iX_i+jX_j + kX_k$ and $Y=Y_r+iY_i+jY_j+kY_k$ \cite{Ogunfunmi2015} (see Fig. \ref{fig:MSE_Corr}). The factor of 4 is introduced to normalize the effect of the four components in the quaternion as opposed to a single real component. In the complex case, the factor of 2 would be included for normalization.
\\
\section{Quaternion Recurrent Neural Network with RTRL and Maximum Correntropy Criterion}
\label{QRNN_algo_MCC}
\begin{figure}
    \centering
    \includegraphics[width=\linewidth]{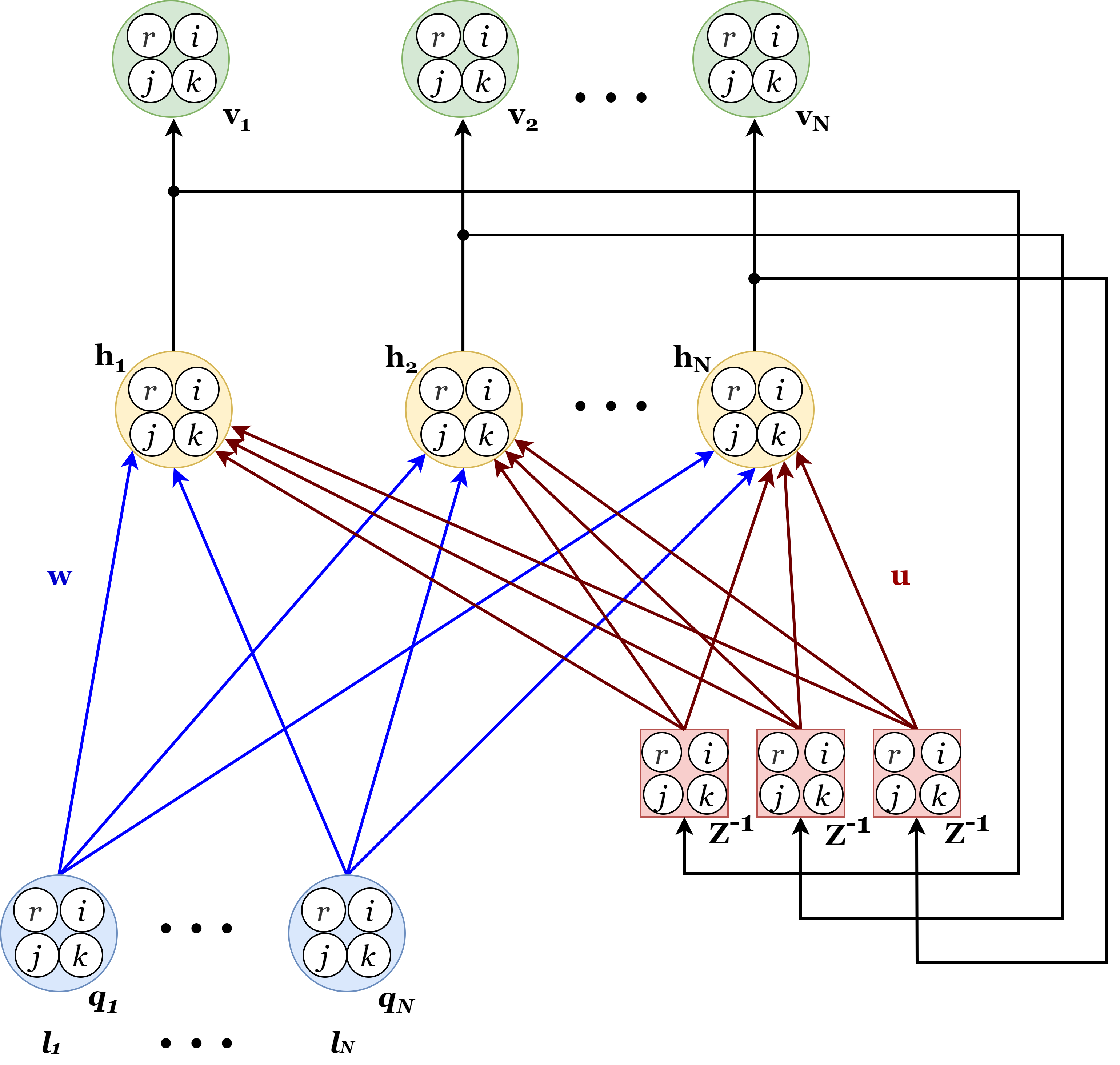}
    \caption{A general architecture for QRNN with RTRL.}
    \label{fig:QRNN}
\end{figure}

We next introduce the QRNN equipped with RTRL and MCC; this is achieved based on the GHR calculus. An illustration of a general architecture for QRNN with RTRL is shown in Fig. \ref{fig:QRNN}.

\subsection{Forward pass}
\label{sec:forward_pass_MCC}
Let $\textbf{h}_a^{(l)}(n)$ denote the hidden state of the $a$th neuron in the $l$th layer at time $n$, defined as
\begin{equation}
\textbf{h}_a^{(l)}(n) = \Phi\,\Bigl(\textbf{f}_a^{(l)}(n)\Bigr)\,,
\end{equation}
where
\begin{equation}
\textbf{f}_a^{(l)}(n) = \sum_{b=1}^{A_{l-1}}  \textbf{u}_{ab}^{(l)} \pmb{\otimes} \textbf{h}_b^{(l-1)}(n-1) + \textbf{w}_{ab}^{(l)} \pmb{\otimes} \textbf{v}_b^{\,(l-1)}(n) + \textbf{b}_a^{(l)} \, .
\end{equation}
Here, $\textbf{u}_{ab}^{(l)}$ represents the vector of recurrent quaternion weights, $\Phi$ is the quaternion split activation function (see Appendix \ref{sec:Appendix_split_activ}), $\textbf{v}_b^{(l-1)}(n)$ represents the output of the $b$-th neuron in layer $l-1$ at time $n$, $\textbf{h}_b^{(l-1)}(n-1)$ is the hidden state of the $b$-th neuron in layer $l-1$ at time $n-1$, and $\pmb{\otimes}$ is the Hamilton product.

\subsection{Correntropy cost function} 
Denote the error vector by $\mathbf{e}(n) = \mathbf{d}(n) - \mathbf{h}(n)$. Given the quaternion Gaussian kernel in (\ref{eq:MCC_quat_kernel}), the correntropy between $\textbf{\textbf{h}}(n)$ and the desired output $\mathbf{\textbf{d}}(n)$ becomes
\begin{equation}
    V_{\sigma} \bigl(\textbf{e}(n) \bigr) = E\Bigl[\kappa_{\sigma} \bigl(\textbf{e}(n)\bigr) \Bigr]\,,
\end{equation}
where
\begin{equation}
    \kappa_{\sigma}\bigl(\textbf{e}(n) \bigr) = \frac{4}{\sqrt{2\pi}\sigma} \exp\biggl[-\frac{1}{2\sigma^2} | \textbf{e}(n)|^2 \biggr]\,.
\end{equation}
Usually, the joint probability density function is unknown and only a finite set of samples is available. To estimate the correntropy, the expectation can be approximated based on 
\begin{equation} \hat{V}_{N,\sigma } \bigl(\textbf{d}(n),\textbf{h}(n)\bigr)=\frac{1}{N}\sum_{n=1}^N {\kappa_\sigma \bigl(\textbf{d}(n) - \textbf{h}(n)\bigr)}. 
\end{equation} 
The correntropy should account for the even moments between $\textbf{d}(n)$ and $\textbf{h}(n)$, especially the magnitude $|\textbf{d}(n) - \textbf{h}(n) |$. As $\sigma$ increases, higher-order terms should diminish, making the second-order term dominant. This facilitates using a gradient based method to maximize correntropy.

As we aim to use gradient descent, instead of maximising $\hat{V}_{N,\sigma } (\textbf{d}(n),\textbf{h}(n))$, the objective of the network training is to minimise the loss function $-\hat{V}_{N,\sigma} (\textbf{d}(n),\textbf{h}(n))$, expressed as
\begin{equation}
    J(n) = - \frac{1}{N} \sum_{n=1}^{N}\kappa_\sigma \bigl(\textbf{e}(n)\bigr) \,,
\end{equation}
where $n$ is the current time index and $N$ is the total number of time steps. Using the correntropy with a Gaussian kernel function as a cost function gives our problem kernel Hilbert space attributes, which differs from the linear MSE that only considers the Euclidean distance of errors. Essentially, the correntropy represents a weighted sum of error distances, while the kernel function modifies the influence of each error by mapping it from the input space to the reproducing kernel Hilbert space.
\subsection{Backpropagation} 
We now derive the quaternion real-time recurrent learning algorithm for the QRNN. The quaternion gradient of the correntropy cost function can be expressed as 
\begin{equation}
\nabla_{\textbf{q}^*}\,J(n) =  \biggl( \frac{\partial J(n)}{\partial \mathbf{q}^*}\biggr)^T =  \biggl( \frac{\partial J(n)}{\partial \mathbf{q}}\biggr)^H \,.
\end{equation}
To find $\frac{\partial J(n)}{\partial \textbf{q}}$, we use the GHR chain rule \cite{Xu2015_GHR} to obtain
\begin{equation}
    \frac{\partial J(n)}{\partial \textbf{q}} = - \frac{1}{N} \sum_{n=1}^{N} \: \sum_{\mu \in \{ 1,i,j,k\}} \frac{\partial \kappa_\sigma (\textbf{e}(n))}{\partial \textbf{e}^\mu(n)}\frac{\partial \textbf{e}^\mu(n)}{\partial \textbf{q}} \,.
\end{equation}
The first term  $\frac{\partial \kappa_\sigma(\textbf{e}(n))}{\partial \textbf{e}^\mu(n)}$ can be evaluated as
\begin{equation}\label{eq:1stPartialDerivative}
    \frac{\partial \kappa_\sigma(\textbf{e}(n))}{\partial \textbf{e}^\mu(n)} = \sum_{\mu \in \{ 1,i,j,k\}} - \frac{4}{\sqrt{2\pi}\sigma^3} \exp \biggl[ - \frac{1}{2\sigma^2} |\textbf{e}(n)|^2 \biggr] \frac{\partial |\textbf{e}(n)|^2}{\partial \textbf{e}^\mu(n)}.
\end{equation}
Upon substituting (\ref{eq:1stPartialDerivative}) into the formula for $\frac{\partial J(n)}{\partial \textbf{q}}$, we obtain
\begin{equation}
\begin{split}
    \frac{\partial J(n)}{\partial \textbf{q}} & = \frac{4}{N\sqrt{2\pi}\sigma^3} \sum_{n=1}^{N} \: \sum_{\mu \in \{ 1,i,j,k\}} \exp \biggl[ - \frac{1}{2\sigma^2} |\textbf{e}(n)|^2 \biggr] \\& \quad \quad \frac{\partial |\textbf{e}(n)|^2}{\partial \textbf{e}^\mu(n)}\frac{\partial \textbf{e}^\mu(n)}{\partial \textbf{q}} \,.
\end{split}
\end{equation}
Then, upon employing the GHR chain rule \cite{Xu2015_GHR}, the derivative of $|\textbf{e}(n)|^2$ can be calculated as
\begin{equation}
    \frac{\partial |\textbf{e}(n)|^2}{\partial \mathbf{q}} = \sum_{\mu \in \{1,i,j,k\}} \frac{\partial |\textbf{e}(n)|^2}{\partial \textbf{e}^\mu(n)} \frac{\partial \textbf{e}^\mu(n)}{\partial \mathbf{q}} \,,
\end{equation}
whereby the GHR product rule gives
\begin{gather}
\begin{split}
    \frac{\partial |\textbf{e}(n)|^2}{\partial \textbf{e}^\mu(n)} & = \frac{\partial (\textbf{e}^*(n)\textbf{e}(n))}{\partial \textbf{e}^{\mu}(n)} = \textbf{e}^*(n) \frac{\partial \textbf{e}(n)}{\partial \textbf{e}^{\mu}(n)} + \frac{\partial \textbf{e}^*(n)}{\partial \textbf{e}^{\textbf{e}(n)\mu}(n)}\textbf{e}(n) \\& = \frac{1}{2}\textbf{e}^\mu(n) \,.\raisetag{1\baselineskip}
\end{split}
\end{gather}
Given that $\textbf{e}(n) = \textbf{d}(n) - \textbf{h}(n)$, we obtain
\begin{equation}
    \frac{\partial \textbf{e}^\mu(n)}{\partial \textbf{q}} = -\Bigl( J_{\textbf{q}^\mu(n)} \Bigr)^\mu \,,
\end{equation}
where $J_{\textbf{q}^{\mu}(n)} \triangleq \frac{\partial \textbf{h}(n)}{\partial \textbf{q}^{\mu}}$ is the Jacobian matrix of $\textbf{h}(n)$. Thus,
\begin{equation}
     \frac{\partial |\textbf{e}(n)|^2}{\partial \mathbf{q}} = - \sum_{\mu \in \{1,i,j,k\}} \frac{1}{2}\textbf{e}^\mu (n)\Bigl( J_{\textbf{q}^\mu(n)} \Bigr)^\mu \,,
\end{equation}
which allows us to arrive at
\begin{equation}
\begin{split}
    \frac{\partial J(n)}{\partial \textbf{q}} & = \frac{4}{N\sqrt{2\pi}\sigma^3} \sum_{n=1}^{N} \: \sum_{\mu \in \{ 1,i,j,k\}} \exp \biggl[ - \frac{1}{2\sigma^2} |\textbf{e}(n)|^2 \biggr] \\ & \quad \quad \biggl( - \frac{1}{2}\textbf{e}^\mu(n) \Bigl( J_{\textbf{q}^\mu(n)} \Bigr)^\mu \biggr) \,.
\end{split}
\end{equation}
This can be simplified into
\begin{equation}\label{eq:MCC_dJ/dq}
\begin{split}
    \frac{\partial J(n)}{\partial \textbf{q}} &= - \frac{2}{N\sqrt{2\pi}\sigma^3} \sum_{n=1}^{N} \:  \exp \biggl[ - \frac{1}{2\sigma^2} |\textbf{e}(n)|^2 \biggr] \\ & \quad \quad \sum_{\mu \in \{ 1,i,j,k\}}  \Bigl( J_{\textbf{q}^\mu(n)} \textbf{e}(n) \Bigr)^\mu  \,.
    \end{split}
\end{equation}
Finally, to find the quaternion gradient, we take the Hermitian (conjugate transpose) of (\ref{eq:MCC_dJ/dq}), that is
\begin{gather}
\begin{split}
\nabla_{\textbf{q}^*}\,J(n) & =  \biggl( \frac{\partial J(n)}{\partial \mathbf{q}}\biggr)^H \\&
     = \biggl( - \frac{2}{N\sqrt{2\pi}\sigma^3} \sum_{n=1}^{N} \:  \exp \biggl[ - \frac{1}{2\sigma^2} |\textbf{e}(n)|^2 \biggr] \\ & \quad \quad \sum_{\mu \in \{ 1,i,j,k\}} \Bigl( J_{\textbf{q}^\mu(n)}\textbf{e}(n) \Bigr)^\mu \biggr)^H \,. \raisetag{1\baselineskip}
\end{split} 
\end{gather}
Denote the quaternion error terms at time $n$ as $\delta^{(l)}(n)$, that is
\begin{gather}
  \delta^{(l)}(n) =
  \begin{cases}
  \textbf{e}(n) = \bigl(\textbf{d}(n)-\textbf{h}^{(l)}(n)\bigr)\,, & \parbox[t]{5.5cm}{$l = L\,, \\ n = N$} \\
    \!\begin{aligned}
    \begin{split}
       & \bar{\Phi}\bigl(\textbf{f}^{\,(l)}(n)\bigr) \odot \biggl( \exp \biggl[ - \frac{1}{2\sigma^2} \left|\delta^{(l+1)}(n)\right|^2 \biggr] \\
       & \quad \biggl[ \Bigl[\bigl(\textbf{W}^{(l+1)}\bigr)^{\mu} \times \bigl(\delta^{(l+1)}(n)\bigr)^{\mu} \Bigr]^H \\
       & \quad + \Bigl[ \bigl(\textbf{U}^{(l)}\bigr)^{\mu} \times \bigl(\delta^{(l)}(n+1)\bigr)^{\mu} \Bigr]^H  \biggr] \biggr) \,,
    \end{split}
    \end{aligned}           & \parbox[t]{5.5cm}{other-\\wise} 
  \end{cases}\raisetag{1\baselineskip}
\end{gather}
with $\bar{\Phi}$ as the derivative of the activation $\Phi$, $\odot$ as the Hadamard product, and $(\textbf{W}^{(l+1)})^H$ and $(\textbf{U}^{(l)})^H$ as the Hermitians of the weight matrices for the $(l+1)$th and the $l$th layer, respectively.
Then, the weight and bias updates for the output layer are
\begin{equation}
\textbf{W}^{(l)} = \textbf{W}^{(l)} + \alpha \sum_{n=1}^{N} \bigl(\delta^{(l)}(n) \bigr) \times \bigl(\textbf{v}^{(l-1)}(n)\bigr)^H \, ,
\end{equation}
\begin{equation}
\textbf{b}^{(l)} = \textbf{b}^{(l)} + \alpha \sum_{n=1}^{N} \delta^{(l)}(n)\,.
\end{equation}
The weight and bias updates for the hidden layers are
\begin{gather}
\textbf{W}^{(l)} = \textbf{W}^{(l)} + \alpha \,\sum_{n=1}^{N} \sum_{\mu \in \{1, i, j, k \}} \bigl( \delta^{(l)}(n) \bigr) \times \bigl(\textbf{v}^{(l-1)}(n)\bigr)^H \, , \raisetag{0.8\baselineskip}
\end{gather}
\begin{equation}
\textbf{U}^{(l)} = \textbf{U}^{(l)} + \alpha \,\sum_{n=2}^{N} \sum_{\mu \in \{1, i, j, k \}} \bigl( \delta^{(l)}(n) \bigr) \times \bigl(\textbf{h}^{(l)}(n)\bigr)^H , \,
\end{equation}
\begin{equation}
\textbf{b}^{(l)} = \textbf{b}^{(l)} + \alpha \,\sum_{n=1}^{N} \sum_{\mu \in \{1, i, j, k \}} \delta^{(l)}(n)  \,.
\end{equation}
Here, $\alpha > 0$ is the learning rate, and $\textbf{U}^{(l)}$ represents the matrix of recurrent quaternion weights for layer $l$.
\\
\section{Simulation results}
\label{sec:typestyle}
\begin{table*}[]
\centering
\renewcommand{\arraystretch}{1.4}
\caption{Forecasting performances of the QRNN RTRL with MCC or MSE loss, RNN RTRL with MCC or MSE loss, QLMS, and LMS. The errors indicate the average of the performance measure over the sequences included in the test dataset for each algorithm. The 95\% mean confidence intervals for the QRNNs and RNNs are shown. The error types presented are the root mean squared error (RMSE, in mm), normalised RMSE (nRMSE, no unit), mean average error (MAE, in mm), and jitter (in mm).}
\begin{tabular}{ccccc}
\hline
\textbf{Error}  & \begin{tabular}[c]{@{}l@{}}\textbf{Prediction method}\end{tabular} & \textbf{All sequences} & \textbf{Regular breathing}           & \textbf{Irregular breathing} \\
\hline

\multirow{6}{*}{RMSE}
    &  \cellcolor{gray!25}{QRNN RTRL w/ MCC} & \cellcolor{gray!25}{\boldmath{$1.486 \pm 0.004$}} & \cellcolor{gray!25}{$1.245 \pm 0.004$} & \cellcolor{gray!25}{\boldmath{$1.683 \pm 0.005$}} \\ 
    &  \begin{tabular}[c]{@{}l@{}}QRNN RTRL w/ MSE\end{tabular} & $1.513 \pm 0.006$ & \boldmath{$1.239 \pm 0.005 $} & $1.741 \pm 0.008$ \\ 
    &  \begin{tabular}[c]{@{}l@{}}RNN RTRL w/ MCC\end{tabular} & $1.691 \pm 0.006$ & $1.437 \pm 0.004$ & $1.896 \pm 0.005$ \\ 
    &  \begin{tabular}[c]{@{}l@{}}RNN RTRL w/ MSE\end{tabular} & $1.709 \pm 0.005$ & $1.421 \pm 0.005 $      & $1.932 \pm 0.006$  \\ 
    &  \begin{tabular}[c]{@{}l@{}}QLMS\end{tabular} & $1.696$ & $1.549$ & $2.004$\\                     
    &  \begin{tabular}[c]{@{}l@{}}LMS\end{tabular}& $1.728$ & $1.573$ & $2.038$ \\
    \hline 

\multirow{6}{*}{nRMSE}    
    & \cellcolor{gray!25}{\begin{tabular}[c]{@{}l@{}}QRNN RTRL w/ MCC\end{tabular}} & \cellcolor{gray!25}{\boldmath{$0.3148 \pm 0.0007$}} & \cellcolor{gray!25}{$0.3115 \pm 0.0006$} & \cellcolor{gray!25}{\boldmath{$0.3392 \pm 0.0008$}} \\ 
    &  \begin{tabular}[c]{@{}l@{}}QRNN RTRL w/ MSE\end{tabular} & $0.3186 \pm 0.0 0 05$ & \boldmath{$0.3092 \pm 0.0004$}  & $0.3473 \pm 0.0011$ \\ 
    &  \begin{tabular}[c]{@{}l@{}}RNN RTRL w/ MCC\end{tabular} & $0.3327 \pm 0.0 0 06$ & $0.3319 \pm 0.0 0 05 $  &  $0.3618 \pm 0.0009$ \\ 
    &  \begin{tabular}[c]{@{}l@{}}RNN RTRL w/ MSE\end{tabular} &  $0.3358 \pm 0.0 0 08$ & $0.3261 \pm 0.0 0 07 $      & $0.3694 \pm 0.0008$  \\ 
    &  \begin{tabular}[c]{@{}l@{}}QLMS\end{tabular} & $0.3461$ & $0.3328$ & $0.4352$\\ 
    &  \begin{tabular}[c]{@{}l@{}}LMS\end{tabular} & $0.3509$ & $0.3391$ & $0.4427$ \\
    \hline 

\multirow{6}{*}{MAE}    
    & \cellcolor{gray!25}{\begin{tabular}[c]{@{}l@{}}QRNN RTRL w/ MCC\end{tabular}}& \cellcolor{gray!25}{\boldmath{$0.752 \pm 0.007$}} & \cellcolor{gray!25}{\boldmath{$0.617 \pm 0.005$}} & \cellcolor{gray!25}{\boldmath{$0.882 \pm 0.009$}} \\ 
    &  \begin{tabular}[c]{@{}l@{}}QRNN RTRL w/ MSE\end{tabular} & $0.771 \pm 0.006$ & $0.625 \pm 0.006$   & $0.918 \pm 0.007$ \\ 
    &  \begin{tabular}[c]{@{}l@{}}RNN RTRL w/ MCC\end{tabular} & $0.854 \pm 0.005$ & $0.701 \pm 0.003$ & $0.971 \pm 0.004$ \\ 
    &  \begin{tabular}[c]{@{}l@{}}RNN RTRL w/ MSE\end{tabular} & $0.863 \pm 0.003$ & $0.713 \pm 0.002$   & $1.002 \pm 0.003$ \\ 
    &  \begin{tabular}[c]{@{}l@{}}QLMS\end{tabular} & $0.935$ & $0.881$ & $1.156$ \\   
    &  \begin{tabular}[c]{@{}l@{}}LMS\end{tabular}& $0.988$ & $0.936$ & $1.214$ \\ 
    \hline 

\multirow{6}{*}{Jitter} 
    &  \cellcolor{gray!25}{\begin{tabular}[c]{@{}l@{}}QRNN RTRL w/ MCC\end{tabular}} & \cellcolor{gray!25}{$0.7083 \pm 0.0013$} & \cellcolor{gray!25}{$0.6217 \pm 0.0012$} & \cellcolor{gray!25}{\boldmath{$0.7741 \pm 0.0014$}} \\ 
    &  \begin{tabular}[c]{@{}l@{}}QRNN RTRL w/ MSE\end{tabular} & \boldmath{$0.6971 \pm 0.0014$} & \boldmath{$0.5948 \pm 0.0014$} & $0.8201 \pm 0.0016$ \\ 
    &  \begin{tabular}[c]{@{}l@{}}RNN RTRL w/ MCC\end{tabular} & $0.7899 \pm 0.0016$ & $0.6932 \pm 0.0014$ & $0.8803 \pm 0.0013$ \\ 
    &  \begin{tabular}[c]{@{}l@{}}RNN RTRL w/ MSE\end{tabular} & $ 0.7862 \pm 0.0014$ & $ 0.6745 \pm 0.0017$    & $0.9021 \pm 0.0012$  \\ 
    &  \begin{tabular}[c]{@{}l@{}}QLMS\end{tabular} & $1.693$ & $1.68$ & $1.726$\\                                        
    &  \begin{tabular}[c]{@{}l@{}}LMS\end{tabular} & $1.697$ & $1.683$ & $1.728$ \\ 
    \hline 
\end{tabular}%
\label{tab:RT_algo_performance}
\end{table*}
\subsection{Motion prediction of chest internal points for lung cancer radiotherapy}
In the context of lung cancer radiotherapy, tracking the position of infrared-reflective markers on the chest is a method used to approximate the location of the tumour. Nonetheless, the precision of radiation delivery in radiotherapy systems is constrained by the inherent latency due to limitations in robotic control. Compensating for delays is crucial to reduce the damage to healthy tissues. Employing RNN with online learning can facilitate predictions that adapt to the dynamic nature of respiratory signals, potentially mitigating this issue \cite{RNN_RTRL_lungRT}. Unlike offline methods, which do not change after initial training, online methods continually update the synaptic weights of the network with each new data input. This continuous updating allows the neural network to adjust to the patient's evolving breathing patterns, offering improved resilience against complex movements. Online learning is particularly advantageous in medical settings, where acquiring extensive training datasets can be challenging. It provides a way to adapt to data not included in the initial training set. Adaptive or dynamic learning techniques have been frequently utilized in radiotherapy. Studies have shown the effectiveness of these approaches compared to static models \cite{Krauss2011,Teo2018,Mafi2020}. One such dynamic method is RTRL \cite{Williams1989}, which has been applied in various systems like the Cyberknife Synchrony system \cite{Mafi2020} and the SyncTraX system \cite{Jiang2019}. It has also been used to track the position of internal points within the chest \cite{RNN_RTRL_lungRT}, demonstrating its broad applicability in medical technology.

While prior studies in the field of respiratory motion prediction primarily concentrated on univariate signal analysis, our simulation focuses on the prediction of 3D marker position, represented as pure quaternions, setting a prediction horizon of two seconds. This is achieved through the use of a QRNN trained with RTRL. We compare its effectiveness, when associated with the MCC loss, against other forecasting algorithms. Furthermore, we conducted predictions for all three markers concurrently, enabling the QRNN to further identify and leverage the interrelationships in their movements.

\subsection{3D marker position data}
The data used for this study was collected from nine instances of 3D positional tracking of three external markers placed on the chest and abdomen of subjects resting on a HexaPOD treatment couch. The tracking was accomplished using an NDI Polaris infrared camera. Each recorded sequence varied in length, ranging from 73 to 320 seconds, and was sampled at a frequency of 10 Hz. The movement trajectories of the markers were recorded in three dimensions: superior-inferior (ranging from 6 mm to 40 mm), left-right (2 mm to 10 mm), and antero-posterior (18 mm to 45 mm). Out of these recordings, five depict normal breathing patterns, while the remaining four capture subjects engaging in activities such as talking or laughing. Further specifics about the public dataset are elaborated in \cite{Krilavicius_dataRT, Pohl2022}. The data from subjects was divided into two groups to assess the robustness of the algorithms.

\subsection{Algorithms \& Training}
We compared the performance of the QRNN equipped with RTRL and the MCC loss function (Sec.~\ref{QRNN_algo_MCC}) against its counterpart that employs the MSE loss function (see Appendix \ref{sec:Appendix:QRNN_algo_MSE}). The performances of the standard RNN with RTRL and MSE loss as described in \cite{RNN_RTRL_lungRT} or MCC loss as defined in (\ref{eq:MCC_real}), as well as the Quaternion Least Mean Square (QLMS) algorithm \cite{Took2009}, and the real-valued Least Mean Square (LMS) algorithm, were also included in the comparison of the prediction methods. Note that the predictions for all three markers were conducted concurrently for the standard RNNs with RTRL. The QRNNs and RNNs were characterized by one hidden layer. The activation function was the hyperbolic tangent function for RNNs, and the split hyperbolic tangent function for QRNNs (see Appendix~\ref{sec:Appendix_split_activ}). The cross-validation metric was the RMSE. The ranges of hyper-parameters for cross-validation with grid search were as follows. For QRNNs and RNNs with RTRL, the learning rate range was $\eta \in \{0.02, 0.05, 0.1, 0.2\}$. The range in the number of hidden units was $h \in \{10, 20, 30, 40, 45\}$ for QRNNs and $h \in \{20, 40, 60, 80, 90\}$ for RNNs. For QLMS and LMS, the parameter range was $\eta \in \{0.0 02, 0.0 05, 0.01,0.02, 0.05, 0.1, 0.2\}$. For all prediction methods, the regressor range defined as the number of time steps was $l \in \{10, 30, 50, 70, 90\}$. There were 50 runs successively performed for cross-validation, and 300 runs for evaluation. Updating QRNNs and RNNs with the gradient rule may induce numerical instability. The gradient norm was therefore clipped to avoid large weight updates, while the optimization method was stochastic gradient descent. The training and validation of these models were conducted in the first minute of each recorded sequence, both for 30 seconds. The four error types, presented in Table \ref{tab:RT_algo_performance}, used to evaluate the prediction performance are implemented following \cite{Pohl2022}.

\subsection{Prediction performance}
The QRNN equipped with RTRL and the MCC loss produced the lowest mean average error, RMSE, and nRMSE averaged over all the sequences as well as in the context of irregular breathing (Table \ref{tab:RT_algo_performance}). The QRNN with RTRL and the MSE loss performed slightly better than its counterpart with the MCC loss for predicting sequences of regular breathing, in terms of the RMSE and nRMSE metrics. The robustness of the QRNN with RTRL and the MCC loss against irregular movements was further confirmed as its nRMSE exhibited an increase of 8.2\% between regular and irregular breathing patterns, which is the smallest among the algorithms assessed. More generally, the QRNNs with RTRL displayed consistently better performance for forecasting the breathing sequences than the standard RNNs with RTRL, QLMS, and LMS algorithms. 
The compact 95\% confidence intervals accompanying the performance metrics presented in Table~\ref{tab:RT_algo_performance} suggest that choosing 300 test runs was adequate to yield precise results. In addition, the jitter was evaluated. It refers to the fluctuation or oscillation amplitude in the predicted signal, and its extent can significantly affect robotic control during treatment. The QRNN with RTRL and MSE displayed a slightly lower jitter overall than its counterpart with the MCC loss. The QRNN with RTRL and MCC exhibited the lowest jitter in the context of irregular breathing, significantly outperforming the other algorithms, including the RNN with RTRL and MCC (0.77 mm against 0.88 mm) and the QLMS and LMS algorithms (0.77 mm against 1.73 mm).

\subsection{Discussion of simulation limits}
The simulation has certain limitations, particularly regarding the quantity and length of the sequences utilized, which are relatively modest. Nonetheless, the dataset used encompasses a broad spectrum of respiratory patterns, including various shifts, drifts, slow and abrupt irregular movements, and both active and calm breathing states. A notable aspect of the QRNNs with RTRL and the MCC or MSE loss function we examined is their ability to perform online learning, which does not require extensive pre-existing data to make precise predictions. This is evident from the high accuracy achieved with just one minute of training data. Based on these factors, we believe our results could be extrapolated to larger datasets. Another strength of the simulations is the open-source dataset we used, which promotes reproducibility of our findings. This contrasts with many prior studies in this field, which often rely on private datasets, complicating comparative analyses. We also addressed challenging scenarios like laughing and talking, which are generally controlled in clinical settings. Analyzing performance under such conditions provides insights into other less predictable events that may occur during treatment, such as yawning, hiccupping, or coughing. The current clinical practice is to halt radiation when such anomalies are detected. By differentiating between regular and irregular breathing patterns, we could objectively assess and measure the resilience of the compared algorithms. Given that almost half of our dataset consisted of irregular breathing sequences, the average numerical error metrics across all nine sequences might be higher than what would be encountered in more standard scenarios. Moreover, the QRNN models do take roughly 50\% longer to train than the RNN models (Table \ref{tab:RT_algo_time}). The QRNN with RTRL and MCC demonstrated an average computation time per time step of 184.5 ms. Note that this computation time is less than the approximate marker position sampling interval, which is around 400 ms.
\begin{table}[]
\caption{Time performance of the QRNN trained with RTRL and MCC loss in comparison with other prediction methods (Ubuntu Linux i7-10700K 3.8GHz CPU NVidia GeForce RTX 3080 GPU 32Gb RAM).}
\small
\renewcommand{\arraystretch}{1.4}
\resizebox{\columnwidth}{!}{%
\begin{tabular}{cc}
\hline
\textbf{Prediction algorithm} & \textbf{Calculation time per time step (in ms)} \\ \hline
\cellcolor{gray!25}{QRNN with RTRL and MCC} &  \cellcolor{gray!25}{184.5} \\ 
QRNN with RTRL and MSE & 182.4 \\ 
RNN with RTRL and MCC & 116.1 \\ 
RNN with RTRL and MSE & 115.7 \\ 
QLMS &  0.507 \\ 
LMS &  0.313 \\ 
\hline
\end{tabular}%
}
\label{tab:RT_algo_time}
\end{table}
\\
\section{Conclusion}
\label{sec:conclusion}
The incorporation of the quaternion algebra into RNNs offers an efficient way to capture and hence make use of the multidimensional structures present in 3D and 4D data. When combined with RTRL, the QRNN model has been demonstrated to adapt and learn intricate multidimensional temporal patterns in real-time. While both MSE and MCC are viable cost functions, the choice depends on the specific application and the nature of the data. Simulations in the context of motion prediction of chest internal points for safer lung cancer radiotherapy have shown that the kernel-based similarity within the MCC helps the QRNN to be consistently more robust to outliers in data. The successful training of QRNNs with RTRL, using just one minute of respiratory data per sequence, have moreover demonstrated the efficacy of dynamic training even with constrained data availability.
\\
\section*{Acknowledgement}
\label{sec:acknowledgements}
Pauline Bourigault was supported by the UKRI CDT in AI for Healthcare http://ai4health.io (Grant No. P/S023283/1).
\\
\bibliographystyle{IEEEtran}
\bibliography{IEEEabrv,refs}

\begin{thebibliography}{10}
\providecommand{\url}[1]{#1}
\csname url@samestyle\endcsname
\providecommand{\newblock}{\relax}
\providecommand{\bibinfo}[2]{#2}
\providecommand{\BIBentrySTDinterwordspacing}{\spaceskip=0pt\relax}
\providecommand{\BIBentryALTinterwordstretchfactor}{4}
\providecommand{\BIBentryALTinterwordspacing}{\spaceskip=\fontdimen2\font plus
\BIBentryALTinterwordstretchfactor\fontdimen3\font minus \fontdimen4\font\relax}
\providecommand{\BIBforeignlanguage}[2]{{%
\expandafter\ifx\csname l@#1\endcsname\relax
\typeout{** WARNING: IEEEtran.bst: No hyphenation pattern has been}%
\typeout{** loaded for the language `#1'. Using the pattern for}%
\typeout{** the default language instead.}%
\else
\language=\csname l@#1\endcsname
\fi
#2}}
\providecommand{\BIBdecl}{\relax}
\BIBdecl

\bibitem{Isokawa2003}
T.~Isokawa, T.~Kusakabe, N.~Matsui, and F.~Peper, \emph{Quaternion Neural Network and Its Application}.\hskip 1em plus 0.5em minus 0.4em\relax Springer Berlin, 2003, p. 318–324.

\bibitem{Parcollet2019_reviewQNN}
T.~Parcollet, M.~Morchid, and G.~Linarès, ``A survey of quaternion neural networks,'' \emph{Artificial Intelligence Review}, vol.~53, no.~4, p. 2957–2982, 2019.

\bibitem{ZhuQuatCNN}
X.~Zhu, Y.~Xu, H.~Xu, and C.~Chen, ``Quaternion convolutional neural networks,'' in \emph{Proceedings of the European Conference on Computer Vision (ECCV)}, 2019.

\bibitem{QuatKnowlGraphEmbed}
S.~Zhang, Y.~Tay, L.~Yao, and Q.~Liu, ``Quaternion knowledge graph embeddings,'' in \emph{Proceedings of the Conference on Neural Information Processing Systems (NeurIPS)}, 2019.

\bibitem{Gaudet2018}
C.~J. Gaudet and A.~S. Maida, ``Deep quaternion networks,'' in \emph{Proceedings of the International Joint Conference on Neural Networks ({IJCNN})}, 2018, pp. 1--8.

\bibitem{Parcollet2019}
T.~Parcollet, M.~Morchid, and G.~Linares, ``Quaternion convolutional neural networks for heterogeneous image processing,'' in \emph{Proceedings of the {IEEE} International Conference on Acoustics, Speech and Signal Processing ({ICASSP})}, 2019, pp. 8514--8518.

\bibitem{Takahashi2022}
K.~Takahashi, E.~Tano, and M.~Hashimoto, ``Feedforward{\textendash}feedback controller based on a trained quaternion neural network using a generalised {HR} calculus with application to trajectory control of a three-link robot manipulator,'' \emph{Machines}, vol.~10, no.~5, p. 333, 2022.

\bibitem{Comminiello2019_3D}
D.~Comminiello, M.~Lella, S.~Scardapane, and A.~Uncini, ``Quaternion convolutional neural networks for detection and localization of 3{D} sound events,'' in \emph{Proceedings of the {IEEE} International Conference on Acoustics, Speech and Signal Processing ({ICASSP})}, 2019, pp. 8533--8537.

\bibitem{Xu2016_Optimization}
D.~Xu, Y.~Xia, and D.~P. Mandic, ``Optimization in quaternion dynamic systems: Gradient, \uppercase{H}essian, and learning algorithms,'' \emph{{IEEE} Transactions on Neural Networks and Learning Systems}, vol.~27, no.~2, pp. 249--261, 2016.

\bibitem{Xu2015_GHR}
D.~Xu, C.~Jahanchahi, C.~C. Took, and D.~P. Mandic, ``Enabling quaternion derivatives: The generalized {HR} calculus,'' \emph{Royal Society Open Science}, vol.~2, no.~8, p. 150255, 2015.

\bibitem{Xu2017}
D.~Xu, L.~Zhang, and H.~Zhang, ``Learning algorithms in quaternion neural networks using \uppercase{GHR} calculus,'' \emph{Neural Network World}, vol.~27, no.~3, pp. 271--282, 2017.

\bibitem{Popa2017}
C.~Popa, ``Learning algorithms for quaternion-valued neural networks,'' \emph{Neural Processing Letters}, vol.~47, no.~3, pp. 949--973, 2017.

\bibitem{Williams1989}
R.~J. Williams and D.~Zipser, ``A learning algorithm for continually running fully recurrent neural networks,'' \emph{Neural Computation}, vol.~1, no.~2, pp. 270--280, 1989.

\bibitem{Mandic2001}
D.~P. Mandic and J.~A. Chambers, \emph{Recurrent Neural Networks for Prediction}.\hskip 1em plus 0.5em minus 0.4em\relax Wiley, 2001.

\bibitem{Parcollet2018QRNN}
T.~Parcollet, M.~Ravanelli, M.~Morchid, G.~Linarès, C.~Trabelsi, R.~De~Mori, and Y.~Bengio, ``Quaternion recurrent neural networks,'' in \emph{Proceedings of the International Conference on Learning Representations (ICLR)}, 2018.

\bibitem{Santamaria2006}
I.~Santamaria, P.~P. Pokharel, and J.~C. Principe, ``Generalized correlation function: Definition, properties, and application to blind equalization,'' \emph{{IEEE} Transactions on Signal Processing}, vol.~54, no.~6, pp. 2187--2197, 2006.

\bibitem{Ward1997}
J.~P. Ward, \emph{Quaternions and Cayley Numbers}.\hskip 1em plus 0.5em minus 0.4em\relax Springer Netherlands, 1997.

\bibitem{Sudbery1979}
A.~Sudbery, ``Quaternionic analysis,'' \emph{Mathematical Proceedings of the Cambridge Philosophical Society}, vol.~85, no.~2, pp. 199--225, 1979.

\bibitem{WeifengLiu2006}
W.~Liu, P.~P. Pokharel, and J.~C. Principe, ``Correntropy: A localized similarity measure,'' in \emph{Proceedings of the International Joint Conference on Neural Networks ({IJCNN})}, 2006, pp. 4919--4924.

\bibitem{Ogunfunmi2015}
T.~Ogunfunmi and T.~Paul, ``The quaternion maximum correntropy algorithm,'' \emph{{IEEE} Transactions on Circuits and Systems {II}: Express Briefs}, vol.~62, no.~6, pp. 598--602, 2015.

\bibitem{RNN_RTRL_lungRT}
M.~Pohl, M.~Uesaka, K.~Demachi, and R.~B. Chhatkuli, ``Prediction of the motion of chest internal points using a recurrent neural network trained with real-time recurrent learning for latency compensation in lung cancer radiotherapy,'' \emph{Computerized Medical Imaging and Graphics}, vol.~91, p. 101941, 2022.

\bibitem{Krauss2011}
A.~Krauss, S.~Nill, and U.~Oelfke, ``The comparative performance of four respiratory motion predictors for real-time tumour tracking,'' \emph{Physics in Medicine and Biology}, vol.~56, no.~16, p. 5303–5317, 2011.

\bibitem{Teo2018}
T.~P. Teo, S.~B. Ahmed, P.~Kawalec, N.~Alayoubi, N.~Bruce, E.~Lyn, and S.~Pistorius, ``Feasibility of predicting tumor motion using online data acquired during treatment and a generalized neural network optimized with offline patient tumor trajectories,'' \emph{Medical Physics}, vol.~45, no.~2, p. 830–845, 2018.

\bibitem{Mafi2020}
M.~Mafi and S.~M. Moghadam, ``Real-time prediction of tumor motion using a dynamic neural network,'' \emph{Medical, Biological Engineering, Computing}, vol.~58, no.~3, p. 529–539, 2020.

\bibitem{Jiang2019}
K.~Jiang, F.~Fujii, and T.~Shiinoki, ``Prediction of lung tumor motion using nonlinear autoregressive model with exogenous input,'' \emph{Physics in Medicine \& Biology}, vol.~64, no.~21, p. 21NT02, 2019.

\bibitem{Krilavicius_dataRT}
T.~Krilavicius, I.~Zliobaite, H.~Simonavicius, and L.~Jarusevicius, ``Predicting respiratory motion for real-time tumour tracking in radiotherapy,'' in \emph{Proceedings of the IEEE 29th International Symposium on Computer-Based Medical Systems (CBMS)}, 2016.

\bibitem{Pohl2022}
M.~Pohl, M.~Uesaka, H.~Takahashi, K.~Demachi, and R.~Bhusal~Chhatkuli, ``Prediction of the position of external markers using a recurrent neural network trained with unbiased online recurrent optimization for safe lung cancer radiotherapy,'' \emph{Computer Methods and Programs in Biomedicine}, vol. 222, p. 106908, 2022.

\bibitem{Took2009}
C.~Took and D.~Mandic, ``The {Q}uaternion {LMS} algorithm for adaptive filtering of hypercomplex processes,'' \emph{IEEE Transactions on Signal Processing}, vol.~57, no.~4, p. 1316–1327, 2009.

\bibitem{Benvenuto1992}
N.~Benvenuto and F.~Piazza, ``On the complex backpropagation algorithm,'' \emph{{IEEE} Transactions on Signal Processing}, vol.~40, no.~4, pp. 967--969, 1992.

\end{thebibliography}

\section{Appendices}
\label{sec:appendices}
\subsection{Quaternion RNN with RTRL and Mean Squared Error}
\label{sec:Appendix:QRNN_algo_MSE}
\subsubsection{Forward pass} The forward pass is defined as in (\ref{sec:forward_pass_MCC}).
\\
\subsubsection{Loss function} The network error at the processing layer is defined by $\mathbf{e}(n) = \mathbf{d}(n) - \textbf{h}^{(L)}(n)$ where $\mathbf{d}(n)$ denotes the desired output. The objective of the network training is to minimize a real-valued mean squared error (MSE) loss function \cite{Xu2016_Optimization}, that is
\begin{equation}
    J(n) = \lVert \mathbf{e}(n) \rVert^2 = \mathbf{e}^H(n)\mathbf{e}(n)
\end{equation}

\subsubsection{Backpropagation} We now derive the quaternion RTRL algorithm for the QRNN. The quaternion gradient of the error function can be expressed as 
\begin{equation}
\begin{split}
\nabla_{\textbf{q}^*}\,J(n) & =  \biggl( \frac{\partial J(n)}{\partial \mathbf{q}^*}\biggr)^T =  \biggl( \frac{\partial J(n)}{\partial \mathbf{q}}\biggr)^H \\& = -\frac{1}{2} \sum_{\mu \in \{1,i,j,k\}} \biggl(\textbf{J}_{\textbf{q}^{\mu}}^{H}(n) \, \textbf{e} (n)\biggr)^{\mu}\, ,
\end{split}
\end{equation}
where $\mathbf{J}_{\mathbf{q}^\mu}(n) \triangleq \frac{\partial \mathbf{h}(n)}{\partial \mathbf{q}^\mu}$ is the Jacobian matrix of $\mathbf{h}(n)$ \cite{Xu2016_Optimization}. 
Denote the quaternion error terms in the recurrent setting at time $n$ as $\delta^{(l)}(n)$. Recall that $\delta^{(l)}(n)$ is essentially the local error term that contributes to the gradient $ \nabla_{\textbf{q}^*}\,J(n)$. These error terms can be calculated recursively as
\begin{equation}
  \delta^{(l)}(n) =
  \begin{cases}
  \textbf{e}(n) = (\textbf{d}(n)-\textbf{h}^{(L)}(n))\,, & \parbox[t]{5.5cm}{$l = L\,, \\ n = N$} \\
    \!\begin{aligned}
       & \bar{\Phi}(\textbf{f}^{\,(l)}(n)) \odot \Bigl[ (\textbf{W}^{(l+1)})^H \times (\delta^{(l+1)}(n)) \\
       & \quad \quad + (\textbf{U}^{(l)})^H \times (\delta^{(l)}(n+1))\Bigr] \,,
    \end{aligned}           & \parbox[t]{5.5cm}{other-\\wise} 
  \end{cases}
\end{equation}
where $\odot$ is the element-wise (Hadamard) product, and $N$ is the total number of time steps.
The weight and bias update rules for the output layer are
\begin{equation}
\textbf{W}^{(l)} = \textbf{W}^{(l)} + \alpha \sum_{n=1}^{N} \delta^{(l)}(n) \times (\textbf{v}^{(l-1)}(n))^H \, ,
\end{equation}
\begin{equation}
\textbf{b}^{(l)} = \textbf{b}^{(l)} + \alpha \sum_{n=1}^{N} \delta^{(l)}(n)\,.
\end{equation}
\noindent The weight and bias update rules for the hidden layers are
\begin{equation}
\textbf{W}^{(l)} = \textbf{W}^{(l)} + \alpha \,\sum_{n=1}^{N} \sum_{\mu \in \{1, i, j, k \}} \biggl( \delta^{(l)}(n) \biggr)^{\mu} \times (\textbf{v}^{(l-1)}(n))^H \, ,
\end{equation}
\begin{equation}
\textbf{U}^{(l)} = \textbf{U}^{(l)} + \alpha \,\sum_{n=2}^{N} \sum_{\mu \in \{1, i, j, k \}} \biggl( \delta^{(l)}(n) \biggr)^{\mu} \times (\textbf{h}^{(l)}(n))^H , \,
\end{equation}
\newpage
\begin{equation}
\textbf{b}^{(l)} = \textbf{b}^{(l)} + \alpha \,\sum_{n=1}^{N} \sum_{\mu \in \{1, i, j, k \}} \biggl( \delta^{(l)}(n) \biggr)^{\mu} \,.
\end{equation}
Here, $\alpha > 0$ is the learning rate, $\textbf{U}^{(l)}$ represents the matrix of recurrent quaternion weights for layer $l$, and $(\cdot)^H$ represents the Hermitian transpose.

\subsection{Split activation function}\label{sec:Appendix_split_activ}
Denote the split activation function as
\begin{equation}\label{eq:3}
    \Phi(\cdot) = \Phi_{\zeta}(\cdot) + \Phi_{\zeta}(\cdot)i + \Phi_{\zeta}(\cdot)j + \Phi_{\zeta}(\cdot)k \,,
\end{equation}
with $\Phi_{\zeta}(\cdot)$ representing any real-valued activation function. To perform backpropagation using split activation functions, \cite{Benvenuto1992} proposed a "pseudo-gradient" update wherein the gradient is computed component-wise. Accordingly, the "pseudo-derivative" of $\Phi(\cdot)$ in (\ref{eq:3}), written as $\bar{\Phi}(\cdot)$, is given by
\begin{equation}
     \bar{\Phi}(\cdot) = \bar{\Phi}_{\zeta}(\cdot) + \bar{\Phi}_{\zeta}(\cdot)i + \bar{\Phi}_{\zeta}(\cdot)j + \bar{\Phi}_{\zeta}(\cdot)k \,.
\end{equation}
On the other hand, the compact GHR derivative of any split activation function $\Phi(\cdot)$ is defined as \cite{Xu2016_Optimization}
\begin{equation}\label{eq:12}
    \begin{split}
        \frac{\partial \Phi(q)}{\partial q} & = \frac{1}{4} \biggl( \frac{\partial \Phi_{\zeta}(q)}{\partial r} - \frac{\partial \Phi_{\zeta}(q)}{\partial x}i - \frac{\partial \Phi_{\zeta}(q)}{\partial y}j - \frac{\partial \Phi_{\zeta}(q)}{\partial z}k \biggr) \\
        & = \frac{1}{4} \Bigl(\bar{\Phi}_{\zeta}(r) + \bar{\Phi}_{\zeta}(x) + \bar{\Phi}_{\zeta}(y) + \bar{\Phi}_{\zeta}(z)\Bigr).
    \end{split}
\end{equation}

\end{document}